\documentclass[conference]{IEEEtran}
\usepackage[top=.75in, left= 0.625in, right = 0.63in, bottom =1in]{geometry}
\IEEEoverridecommandlockouts
\usepackage{cite}
\usepackage{algorithm}
\usepackage{algorithmic}
\usepackage{graphicx}
\usepackage{textcomp}
\usepackage{xcolor}
\usepackage{subcaption}
\usepackage{graphicx} 
\usepackage{amsmath}
\usepackage{amssymb}
\usepackage{amsfonts}
\usepackage{relsize}
\usepackage{mathtools}
\usepackage{amsthm}
\graphicspath{ {./Figures/} }
\def\BibTeX{{\rm B\kern-.05em{\sc i\kern-.025em b}\kern-.08em
    T\kern-.1667em\lower.7ex\hbox{E}\kern-.125emX}}

\newtheorem{definition}{Definition}
\newtheorem{remark}{Remark}
\newtheorem{assumption}{Assumption}
\newtheorem{theorem}{Theorem}
\newtheorem{lemma}{Lemma}

\newcommand{\mathleft}{\@fleqntrue\@mathmargin0pt}
\newcommand{\mathcenter}{\@fleqnfalse}

\DeclareMathOperator*{\maximize}{max}
\DeclareMathOperator*{\minimize}{min}
\DeclareMathOperator*{\subjectTo}{s.t.}

\DeclareMathOperator*{\argmax}{argmax}
\DeclareMathOperator*{\argmin}{argmin}

\DeclareMathOperator{\E}{\mathbb{E}}

\newcommand{\AGENTS}{M}
\newcommand{\agentSet}{\mathcal{M}}
\newcommand{\setX}{\mathcal{X}}
\newcommand{\setY}{\mathcal{Y}}

\newcommand{\vecY}{\boldsymbol{y}}
\newcommand{\graph}{\mathcal{G}}
\newcommand{\neighbors}{\mathcal{N}}
\newcommand{\vertexSet}{\mathcal{V}}
\newcommand{\edgeSet}{\mathcal{E}}
\newcommand{\datasetLoc}{\mathcal{D}}

\newcommand{\sizeSetLoc}{D}

\newcommand{\thetaSpace}{K}
\newcommand{\myFramework}{\emph{BayGo}}

\begin{document}

\title{BayGo: Joint Bayesian Learning and Information-Aware Graph Optimization}

\author{
	\IEEEauthorblockN{
		Tamara AlShammari,
		Sumudu Samarakoon,
		Anis Elgabli,
		and
		Mehdi Bennis
		\\}
	\IEEEauthorblockA{
		\small%
		Centre for Wireless Communication, University of Oulu, Finland\\ Emails: \{Tamara.Alshammari, Sumudu.Samarakoon, Anis.Elgabli, Mehdi.Bennis\}@oulu.fi 
	}
	\thanks{This work was supported  in part by the INFOTECH Project NOOR, in part by the NEGEIN project, by the EU-CHISTERA projects LeadingEdge and CONNECT, the EU-H2020 project under agreement No. 957218 (IntellIoT) and the Academy of Finland projects MISSION and SMARTER.}
}

\maketitle

\begin{abstract}
This article deals with the problem of distributed machine learning, in which agents update their models based on their local datasets, and aggregate the updated models collaboratively and in a fully decentralized manner. In this paper, we tackle the problem of information heterogeneity arising in multi-agent networks where the placement of informative agents plays a crucial role in the learning dynamics. Specifically, we propose \myFramework{}, a novel fully decentralized joint Bayesian learning and graph optimization framework with proven fast convergence over a sparse graph. Under our framework, agents are able to learn and communicate with the most informative agent to their own learning. Unlike prior works, our framework assumes no prior knowledge of the data distribution across agents nor does it assume any knowledge of the true parameter of the system. The proposed alternating minimization based framework ensures global connectivity in a fully decentralized way while minimizing the number of communication links. We theoretically show that by optimizing the proposed objective function, the estimation error of the posterior probability distribution decreases exponentially at each iteration. Via extensive simulations, we show that our framework achieves faster convergence and higher accuracy compared to fully-connected and star topology graphs. 
\end{abstract}

\section{Introduction}
Recently, distributed machine learning at the network edge has received significant attention thanks to its communication-efficiency, low-latency and privacy-preserving benefits \cite{16}. Among these learning techniques, Federated Learning (FL) has been proposed in \cite{1,2,17} where agents train their models privately without sharing their raw data and under the orchestration of a centralized server for model aggregation. The majority of existing works in FL lack robustness and do not generalize well outside of their training distribution. To achieve better generalization, a Bayesian variant of FL was proposed in \cite{5} in which agents learn a posterior probability distribution over their model parameter.

Aggregating agents' posterior updates in a decentralized manner over a given graph while assuming all agents are equally informative has been investigated in~\cite{5,4,7,10}. However, such approaches do not optimize the graph topology, and overlook the fact that the value of information acquired from agents is instrumental in learning due to information heterogeneity across the network. In view of that, the authors in \cite{3} show that the quality of information available to different agents along with the network structure determine the speed of learning. Therein, they show that when agents with most informative observations have higher eigenvector centrality\footnote{Eigenvector centrality of node $i$ is a measure of social influence and it is proportional to agent $i$'s number of neighbors.}, the rate of learning is maximized. Similar conclusions were drawn in \cite{5} where authors consider a single informative agent in the network and showed that the rate of convergence is maximized when the agent with the most informative local observations has the most influence in terms of highest eigenvector centrality on the network. 

In both \cite{5} and \cite{3}, the most informative agent is defined as the agent with the highest local relative entropy which is the Kullback-Leibler (KL) divergence between the likelihood at the true parameter and the likelihood at any other parameter conditioned on the agent's local observations. However, identifying informative agents according to this definition with no prior knowledge of the true parameter is infeasible. Therefore, in this work, we propose a different approach that accounts for information heterogeneity exploiting the KL-divergence between agents' posteriors. Specifically, we define the most informative agent from each agent's perspective as the one that maximizes the KL-divergence. This definition is motivated by the fact that agents learn faster if different pieces of information are aggregated instead of incorporating redundant information from other agents. The optimization problem is based on the alternating minimization approach to alternate between updating the posterior and learning the graph. Moreover, we show that the proposed solution ensures connectedness of the graph and fast convergence.

Our novel framework, \myFramework{}, allows agents to jointly optimize their posterior distributions and their connections on the graph in a fully distributed manner. Our framework exploits the fact that the divergence between agents' posterior distributions is an indication of how different their local datasets are and how different the knowledge they acquired from their past interactions in the network. In other words, the higher the divergence, the more statistically different are agents' local observations and their acquired knowledge. This calls for learning which agent is the most informative among the direct neighbors of a given agent. Thus, the goal is to ensure that each agent communicates with the highest diverged agent among its direct neighbors at any given time. We show theoretically that the error in estimating the posterior distribution is exponentially decaying at each iteration. Finally, our simulation results show that the proposed approach achieves faster convergence and higher accuracy compared to both fully connected and star topologies.

The rest of the paper is organized as follows. In section \ref{system_model}, we introduce the system model and problem formulation. In section \ref{propAlgo}, we describe our  alternating minimization based algorithm to solve the proposed optimization problem. In section \ref{simulations}, we introduce and discuss our simulation results. Finally, we conclude the paper in section \ref{conclusion}.

\textbf{Notation:} We use boldface lowercase symbol for vectors $\boldsymbol{s}$, and boldface uppercase symbol for matrices $\boldsymbol{S}$. In addition, we refer to the Kullback-Leibler (KL) divergence between two probability distributions as  $D_\textrm{KL}(P_r || P^\prime_r)$ such that $(P_r,P^\prime_r) \in \Delta R$ where $\Delta R$ denotes a set of prabability distributions. Moreover, for simplicity, and without loss of generality, we discretize the parameter space $\Theta$ with $\thetaSpace$ representative points and refer to this set as $\Theta_\thetaSpace$. 

\section{System Model and Problem Formulation}
\label{system_model}
Consider a set $\agentSet = \{1, 2, ..., \AGENTS\}$ of $\AGENTS{}$ agents where each agent $i \in \agentSet$ holds a local dataset  $\datasetLoc_{i}=\{(\boldsymbol{X},\boldsymbol{y}) | \boldsymbol{X} \in \setX_i, \boldsymbol{y} \in \setY\}$  of cardinality $\sizeSetLoc_{i}$ where $\setX_i$ is the local instance space at agent $i$, and $\setY$ is the set of all possible labels. Note that the union of all local instance spaces of all agents is contained in the global instance space $\setX$. i.e., $\cup_{i=1}^\AGENTS \setX_{i} \subset \mathcal{X}$. Each agent $i$ generates samples' labels according to a global probabilistic model with distribution $f(y|\boldsymbol{x})$. Agent $i$'s local samples, $\setX_i = \{\boldsymbol{x}_i^{(1)}, \boldsymbol{x}_i^{(2)}, ..., \boldsymbol{x}_i^{(\sizeSetLoc_i)}\}$, are assumed to be independent and identically distributed (i.i.d). 

Each agent $i \in \agentSet$ aims to learn the true parameter of the system $\theta^* \in \Theta$ where $\Theta$ denotes the set of all possible states. It is also assumed that each agent $i$ holds a set of local likelihood functions of the labels $\{l_i(y|\boldsymbol{x},\theta) | y \in \mathcal{Y}, \boldsymbol{x} \in \mathcal{X}_i, \theta \in \Theta\}$. Furthermore, we denote the posterior distribution of agent $i$ over parameter $\theta$ at time $t \geq 0$ by $\boldsymbol{\mu}_{i,t} \in \Delta\Theta$ where $\Delta\Theta$ is a probability distribution over the set $\Theta$. 

\begin{assumption}
\label{pos_prior}
For each agent $i \in \agentSet$, agent $i$'s prior distribution at $t=0$ is $\mu_{i,0}(\theta_k) > 0, \,\,\, \forall \theta_k \in \Theta$ where $\theta_k$ denotes a possible value for parameter $\theta$. 
\end{assumption}
 
Assumption \ref{pos_prior} is necessary to rule out the degenerate case where zero Bayesian prior prevents learning. We next define $\tilde\Theta_i$ as the set of parameters that are observationally equivalent to $\theta^*$ from the agent's $i$ perspective. Formally, $\tilde\Theta_i = \{\theta_k \in \Theta | l_i(y|\boldsymbol{x},\theta_k) =  l_i(y|\boldsymbol{x},\theta^*) \,\, \forall (\boldsymbol{x},y) \in \datasetLoc_i\}$. In addition, we define $\tilde\Theta = \cap_{i=1}^\AGENTS \tilde\Theta_i$ to represent the set of states that are observationally equivalent to $\theta^*$ from all agents' perspective.

\begin{assumption}
\label{learnable}
There exists a parameter $\theta^*$ that is globally identifiable; i.e. $\tilde\Theta =\{\theta^*\}$. 
\end{assumption}
Assumption \ref{learnable} requires that all agents sets' of states that are observationally equivalent to $\theta^*$ have a single element in common. Moreover, we model the interactions between agents as a weighted directed graph $\graph = (\agentSet,\edgeSet)$ which represents an overlay over a physical graph. The edge set $\edgeSet$ contains pairs of connected agents on $\graph$; i.e. pair $(j,i) \in \edgeSet$ if and only if agent $i$ is receiving updates from agent $j$. We define agent $i$'s set of neighors on $\graph$ as $\neighbors_i = \{j \in \vertexSet : (j,i) \in \edgeSet\}$. We assume that each vertex has a self-loop which represents an edge from itself to itself: $(i,i) \in \edgeSet$ for $\forall i \in \agentSet$. 

We use $w_{ij}$ to denote the weight of the directed edge from agent $j$ to agent $i$. Note that $w_{ij} > 0$ if and only if $j \in \neighbors_i$, otherwise $w_{ij} = 0$. Recall that edge weight $w_{ij}$ represents the influence of agent $j$ in learning the posterior distribution of agent $i$. We represent agent's own influence on its learning as the weight of the agent's self-loop $w_{ii}>0$ which reflects the extent of agents' local learning on aggregated posterior distribution. 
It is worth mentioning that the edge weights matrix $W$ is row-stochastic. i.e., $\sum_{j = 1}^\AGENTS w_{ij} = 1$ for all $i \in \agentSet$.
Finally, to ensure that the local information of each agent can be disseminated within the network, we point out the following remark:
\begin{remark}
\label{connectivity}
To ensure convergence to a global model, the graph needs to be B-strongly-connected. That is, there is an integer $B \geq 1$ such that the graph $\left\{\agentSet, \cup_{z=nB}^{(n+1)B-1} \edgeSet_z \right\}$ is strongly-connected for all $n\geq0$. 
\end{remark}
The goal of each agent $i$ is to learn a posterior distribution $\boldsymbol\mu_i$ that makes the predictive distribution $\sum_\Theta l_i(y|\boldsymbol{x},\theta) \mu_i(\theta)$ as close as possible to the true labeling function $f(y|\boldsymbol{x})$. To measure the divergence between both distributions, we use $D_\textrm{KL}\left(f(y|\boldsymbol{x}) || \sum_\Theta l_i(y|\boldsymbol{x},\theta) \mu_i(\theta)\right)$.  
\begin{definition}
\label{optimal_post}Under Assumptions \ref{pos_prior}, \ref{learnable}, and Remark \ref{connectivity}, agent $i \in \agentSet$ asymptotically learns the true parameter $\theta^*$ on a path ${(\boldsymbol{X}^{(t)},\boldsymbol{y}^{(t)})}_{t=1}^\infty$ if along that path $\lim_{t \rightarrow \infty} \mu_{i,t}(\theta^*) = 1$ \cite{11}.
\end{definition}
Definition \ref{optimal_post} implies that the true labeling function is equivalent to $l(y|\boldsymbol{x},\theta^*)$ since the optimal posterior distribution takes value one at $\theta^*$ and zero elsewhere. Thus, the KL-divergence between the true labeling distribution and predictive distribution can be rewritten as $D_\textrm{KL}\left(l_i(y|\boldsymbol{x},\theta^*) || l_i(y|\boldsymbol{x},\theta)\right)$ which represents the agents' local relative entropy. 

Since all agents collaborate on the graph to learn the global model, any agent $i \in \agentSet$ can learn more efficiently if it assigns higher edge weights $w_{ij}$ to agents with lower KL divergence between their likelihood functions and the true generating likelihood. To this end, we cast the following joint optimization problem,
%
\begin{subequations}\label{eqn:master_problem}
\begin{eqnarray}
	\label{start_obj2}
	\displaystyle\minimize_{\theta \in \Theta, W}
	&&
	\textstyle \sum\limits_{i = 1}^\AGENTS \sum\limits_{j = 1}^\AGENTS w_{ij} D_\textrm{KL}\Big(l_j(y|\boldsymbol{x},\theta^*) || l_j(y|\boldsymbol{x},\theta) \Big) \\
	\label{const1}
	\subjectTo 
	&& \textstyle \sum_{j=1}^\AGENTS w_{ij}=1, w_{ii}\geq \delta \qquad \forall i \in \agentSet,\\
	&& w_{ij}\geq 0 \qquad \forall j \in \mathcal{S}_i, \, j\ne i,\\
	&& w_{ij}=0 \qquad \forall j \notin \mathcal{S}_i,  \, j\ne i, \label{const4}
\end{eqnarray}
\end{subequations}
where $\delta$ is a strictly positive constant, and $\mathcal{S}_i$ is agent $i$'s set of physical direct neighbors.
The above objective is bi-convex with respect to $\theta$ and $W$. Next, we introduce our algorithm to solve the proposed optimization problem defined in \eqref{eqn:master_problem}.

\vspace{+.1cm}
\section{Distributed Learning via Alternating Minimization}
\label{propAlgo}
In this section we describe our alternating minimization based approach to solve the proposed problem defined in \eqref{eqn:master_problem}.
We alternate between updating the agents' posterior distributions given the current social graph and updating the graph given the current posterior distributions.
\subsection{Bayesian Parameter Estimation}
In what follows, we describe our decentralized algorithm for learning the posterior distribution over the model parameter for a given graph. We begin by introducing our first lemma.
\begin{lemma}
\label{lemma1}
The problem of minimizing the KL divergence $D_\textrm{KL}$ between two distributions w.r.t $\theta$ is equivalent to the problem of maximizing the expectation of the logarithm of one distribution with respect to the other one. i.e.,
%
\begin{align}
	\label{kl_mle}
	\min_{\theta \in \Theta} D_\textrm{KL}\left(l_j^* || l_j \right) 
	&= \min_{\theta \in \Theta}\{\E_{l_j^*} \left(\log l_j^*\right) -  \E_{l_j^*} \left(\log l_j\right)\}\\
	&= \max_{\theta \in \Theta} \E_{l_j^*} \left(\log l_j\right),
\end{align}
where $l_j = l_j(y|\boldsymbol{x},\theta)$, $l_j^*=l_j(y|\boldsymbol{x},\theta^*)$, and $\E_{l_j^*}$ is the expectation with respect to $l_j(y|\boldsymbol{x},\theta^*)$. The term $\E_{l_j^*} \left(\log l_j^*\right) $ in \eqref{kl_mle} is ignored since it is not a function of the estimated parameter $\theta$.  
\end{lemma}

Using Lemma \ref{lemma1}, and assuming a given graph, equation \eqref{start_obj2} can be recast in terms of the model parameter as follows:
\begin{equation}
\label{mle}
\textstyle 
\maximize_{\theta \in \Theta} \quad
\sum_{i = 1}^\AGENTS \sum_{j = 1}^\AGENTS w_{ij} \E_{l_j^*} \Big(\log ( l_j(y|\boldsymbol{x},\theta)\Big).
\end{equation}


The Maximum Likelihood Estimation (MLE) problem presented above can be casted as an optimization problem over the posterior distribution vector $\boldsymbol{\mu}$ by reformulating it as an inner product of the posterior vector $\boldsymbol{\mu}$ and expectation of log likelihood \cite{11} as follows:
\begin{equation}
\label{joint_obj}
\textstyle \maximize_{\boldsymbol\mu \in \Delta\Theta} \boldsymbol{\mu}^T 
\sum_{i = 1}^\AGENTS \sum_{j = 1}^\AGENTS w_{ij} \E_{l_j^*} \Big(\log ( l_j(y|\boldsymbol{x},\theta)\Big).
\end{equation}

The equivalence of \eqref{mle} and \eqref{joint_obj} follows immediately from Assumption \ref{learnable} and Definition \ref{optimal_post}, that is, finding the true parameter $\theta^*$ that maximizes the likelihood is equivalent to finding the posterior distribution that gives value one at $\theta^*$ and zero elsewhere.

The major challenge in optimizing the objective in \eqref{joint_obj} lies in the fact that $\E_{l_j^*}(\cdot)$ is unknown which means that the true gradient of the objective cannot be computed. A common approach to tackle the objective in \eqref{joint_obj} is to consider the empirical average as the cost function, and solve the online stochastic learning problem. Particularly, we compute the stochastic gradient which replaces the true gradient at the given update time, and we project it on the feasible set while regularizing the projection using a proximal function enforcing it not to oscillate wildly \cite{10}\cite{11}.

To derive the Bayesian parameter estimation from this setup at iteration $t$, the proximal function needs to be the KL-divergence from the prior distribution (i.e. the distribution at $t-1$). To this end, since all agents' local observations are independent, the per-agent objective in \eqref{joint_obj} can be casted as follows:
%
\begin{equation}
	\label{mirror}
	\boldsymbol{\mu}_{i,t} 
	= \textstyle
	\argmin\limits_{\boldsymbol{b} \in \Delta \Theta} \Big\{ 
	-\boldsymbol{b}_i^T \boldsymbol{g}_{i,t} + 
	\sum\limits_{j=1}^\AGENTS \frac{w_{ij}}{\alpha_t}   D_\textrm{KL}(\boldsymbol{b}_i||\boldsymbol{\mu}_{j,t-1})
	\Big\},
\end{equation}
%
where $\boldsymbol{g}_{i,t}=\sum_{j = 1}^\AGENTS w_{ij} \log ( l_j(\boldsymbol{y}^t|\boldsymbol{X}^t,\theta))$ is a noisy realization of the gradient of the objective presented in \eqref{joint_obj} at time $t$, and $\alpha_t >0$ is the step size. We point out that $(\boldsymbol{X}^t,\boldsymbol{y}^t)$ represents the mini-batch of observations that agent $i$ draws from its local dataset at time $t$. Since $D_\textrm{KL}(p||q)= \sum_{k=1}^K p_k \log\frac{p_k}{q_k}$, and by letting $\alpha_t = 1$ for all $t$, the optimization problem in \eqref{mirror} can be recast as follows:
%
\begin{subequations}\label{eqn:interm_problem}
\begin{eqnarray}
	\label{post_obj}
	\minimize_{\boldsymbol{b} \in \Delta \Theta} 
	&& \textstyle -\boldsymbol{b}_i^T \boldsymbol{g}_{i,t} + \sum\limits_{j=1}^\AGENTS w_{ij} \sum\limits_{k=1}^\thetaSpace [b_{i}]_{k} \log \frac{[b_{i}]_{k}}{\mu_{j,t-1}(\theta_k)} \\
	\subjectTo 
	&& [b_i]_{k} \geq 0, \quad \textstyle \sum_{k=1}^\thetaSpace [b_i]_{k} = 1.
\end{eqnarray}
\end{subequations}
Leaving the positivity constraint implicit, we can write \eqref{post_obj} as the maximization of the following lagrangian, 
\begin{multline}
		\label{post_lag}
		L_i(\boldsymbol{b},\lambda) = 
		\lambda(\boldsymbol{b}_i^T \boldsymbol{1}-1) + \boldsymbol{b}_i^T \boldsymbol{g}_{i,t} \\
		- \textstyle \sum_{j=1}^\AGENTS w_{ij} \sum_{k=1}^\thetaSpace [b_{i}]_{k} \log \frac{[b_{i}]_{k}}{\mu_{j,t-1}(\theta_k)},
\end{multline}
where $\boldsymbol{1}$ is vector of all ones. 
The condition for the stationary point is,
%
\begin{equation}
		\label{post_update}
		\mu_{i,t}(\theta_k) 
		=  \frac {\exp\left(\sum_{j=1}^\AGENTS w_{ij} \log \tilde\mu_{j,t}(\theta_k)\right)}
		{\sum_{q=1}^K \exp\left(\sum_{j=1}^\AGENTS w_{ij} \log \tilde\mu_{j,t}(\theta_q)\right)},
\end{equation}
where $\tilde\mu_{i,t}(\theta_k)$ denotes the agent's local update of the posterior distribution, which is defined as
%
\begin{equation}
	\label{private_post}
	\tilde\mu_{i,t}(\theta_k) 
	= \frac{l_i(\boldsymbol{y}^t_i|\boldsymbol{X}^t_i,\theta_k) \mu_{i,t-1}(\theta_k)}{\sum_{q=1}^K l_i(\boldsymbol{y}^t_i|\boldsymbol{X}^t_i,\theta_q) \mu_{i,t-1}(\theta_q)}.
\end{equation}
%
\subsection{Decentralized Optimization of the Graph}
\label{graph}

\begin{figure*}
	\centering
	\includegraphics[trim=0.1in 0.1in 0.1in 0in, clip,  width=0.9\textwidth]{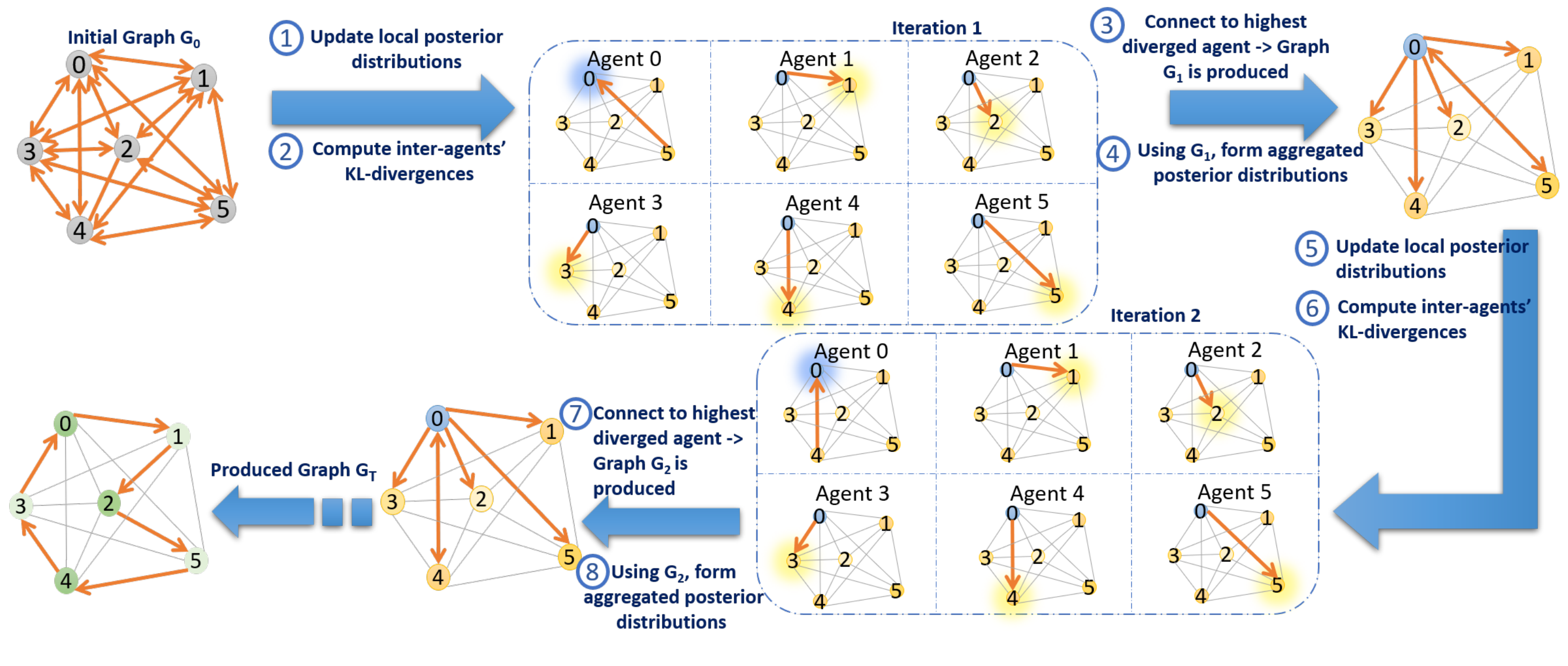}
	\caption{An illustrative example describing the flow of BayGo algorithm}
	\label{algorithm}
\end{figure*}

In this section, we propose a decentralized algorithm for optimizing the graph given the current posterior distributions. The objective in \eqref{start_obj2} implies that agents learn better if they assign higher weights to neighbors with local observations that can distinguish between the true parameter $\theta^*$ and any other parameter $\theta \in \Theta \setminus \Theta^*$. Nevertheless, this objective is infeasible to solve since agents have no prior knowledge of the true parameter $\theta^*$. Mathematically, the first term on the right hand side of \eqref{kl_mle} which represents the likelihood at unknown $\theta^*$ can no longer be ignored because $w_{ij}$ is multiplied by both terms of the KL-divergence. Therefore, unlike prior works, we propose a relaxed problem that does not require any prior knowledge of $\theta^*$ defined as follows:
%
%
%
\begin{subequations}\label{eqn:modified_problem}
\begin{eqnarray}
	\label{opt_w_obj}
	\maximize_{W} 
	&& \textstyle F(W) = \sum_{i = 1}^\AGENTS \sum_{j = 1}^\AGENTS w_{ij} D_\textrm{KL}(\tilde{\boldsymbol{\mu}}_i||\tilde{\boldsymbol{\mu}}_j) \hphantom{text}\\
	\subjectTo 
	&& \eqref{const1}-\eqref{const4}.
\end{eqnarray}
\end{subequations}
The relaxed objective exploits the KL-divergence between agents' posterior distributions (i.e. inter-agents relative entropy) to capture the difference between agents' local observations and the difference in their acquired knowledge from their past interactions in the network. Intuitively, from each agent's perspective, the higher the KL-divergence, the more valuable the relative posterior distribution to the agent's own learning.

The optimization problem \eqref{eqn:modified_problem} is a standard linear programming (LP) problem. 
Hence, the optimal solution for any agent $i \in \agentSet$ at any given time $t$ is
\begin{equation}
w_{ij}^{(t)} =
    \begin{cases}
      \delta, & j = i\\
      1 - \delta, & j = \argmax_j  D_\textrm{KL}(\tilde{\boldsymbol{\mu}}_{i,t}||\tilde{\boldsymbol{\mu}}_{j,t})\\
      0, & \text{otherwise}.
    \end{cases}   
\end{equation}

The solution gives rise to a sparse graph where each agent selects exactly one neighbor that is the most valuable to its learning at time $t$. Not only this solution achieves communication efficiency, but it also outperforms the fully-connected topology in terms of accuracy and rate of convergence (as will be shown in Fig. \ref{fig:Ng1}). This is due to the fact that agents concentrate their mixing weights on posteriors distributions that are most informative to their own learning rather than diluting the weights over many redundant posterior updates.

Interestingly, although we do not enforce any connectivity constraints on the graph, the resulting graph is always globally connected. Fig. \ref{algorithm} shows an illustrative example that describes the flow of our algorithm. In this example, we are considering the extreme case of the existence of a single informative agent in the system (agent 0). At first, agents have no prior knowledge of data distribution across the network. After the first update of the local posterior distributions, agents compute the KL-divergences between their updates, and each agent connects to the highest diverged agent from its own posterior distribution. Particularly, from each agent's perspective, the highest diverged agent is the agent with the very different shade. At iteration $1$, all agents connect to agent $0$ since it is the most diverged agent from their perspective. Nevertheless, from agent $0$ perspective, all agents are comparably diverged so it picks one at random (e.g. agent $5$) and it incorporates agent $5$ posterior distribution in its own posterior distribution. This means that agent's $0$ posterior distribution becomes closer to agent $5$ posterior distribution which means that their KL-divergence probably will not be the highest in subsequent iterations. Thus, at iteration $2$, all agents are still connected to agent $0$ because it is still the most diverged one, though agent $0$ selects a different agent (e.g. agent $4$), and so on. As a result, the union of graphs after $B$ steps is strongly-connected. Eventually, all agents' posteriors will gradually converge (hence, the green shade) and the neighbor selection reduces to a random selection of one agent. 

The details of BayGo are summarized in Algorithm \ref{alg:algorithm}. Now, we are ready to introduce our main theorem which states that by maximizing the objective function in \eqref{opt_w_obj}, the agents' belief of any wrong hypothesis at any given time is exponentially reduced at iteration $t$.

\begin{theorem}
Under Assumptions \ref{pos_prior}, \ref{learnable}, and Remark \ref{connectivity}, using our joint optimization framework for any arbitrary small $\epsilon > 0$, the following condition holds
\begin{equation}
	\left|\mu_{i,t}(\theta^*)-1\right| \leq (\thetaSpace - 1) e^{-\left(K(\Theta) - \epsilon\right)}
\end{equation}
where
$$K(\Theta):= 
\textstyle
\min\limits_{\theta_k \in \Theta \setminus \Theta_K} 
\left(\sum\limits_{j=1,j\ne i}^\AGENTS w_{ij}^{(t)} D_\textrm{KL}(\tilde{\mu}_{i,t}(\theta_k)||\tilde{\mu}_{j,t}(\theta_k))\right).$$
\end{theorem}

Due to space limitations, the proof is omitted. Intuitively, agents' collaboration with the most informative agent to their learning leads to faster learning of the true parameter $\theta^*$. In other words, aggregating the valuable information at any given time leads to faster convergence.

\begin{algorithm}[!b]
\caption{\myFramework{}} 
\hspace*{\algorithmicindent} \textbf{Inputs:} $\boldsymbol{\mu}_{i,0} \in \Delta\Theta$ with $\boldsymbol{\mu}_{i,0} > 0$ for all $i \in \agentSet$, strongly-connected physical graph, $\delta$\\
\hspace*{\algorithmicindent} \textbf{Outputs:} $\boldsymbol{\mu}_{i,t}$ for all $i \in \agentSet$\\
\vspace{-.5cm}
\begin{algorithmic}[1]
\FOR{$t = 1$ \TO $T$}
	\FORALL{$i \in \agentSet$, in parallel}
		\STATE Draw a batch of samples $(\boldsymbol{X}_i^{(t)},\vecY^{(t)})$.
		\STATE Form $\tilde{\boldsymbol{\mu}}_{i,t}$ via \eqref{private_post}. 
		\STATE Send $\tilde{\boldsymbol{\mu}}_{i,t}$ to neighbor $z$ for which $z \in \mathcal{S}_j$, and receive $\tilde{\boldsymbol{\mu}}_{j,t}$ from neighbor $j \in \mathcal{S}_i$.
		\STATE Calculate KL-divergence between $\tilde{\boldsymbol{\mu}}_{i,t}$ and $\tilde{\boldsymbol{\mu}}_{j,t}$, $\forall j \in \mathcal{S}_i$.
		\STATE Select the neighbor with the highest KL-divergence (denoted as agent $s$).
		\STATE Update $w_{is} = 1 - \delta$, and all $w_{ij} = 0, \forall j \in \agentSet \setminus \{i,s\}$.
		\STATE Update $\boldsymbol{\mu}_{i,t}$ via \eqref{post_update}.
	\ENDFOR
\ENDFOR
\end{algorithmic}
\label{alg:algorithm}
\end{algorithm}
\section{Numerical Evaluation}
\label{simulations}
In this section, we evaluate our Bayesian learning framework on a linear regression task with a Gaussian prior over $\theta$ with zero mean vector and covariance matrix given by $diag[0.5,0.5]$. We assume a network of $12$ agents, and we use the bodyfat database \cite{15} where all agents have observations of abdomen feature $x$ to predict bodyfat percentage. Nevertheless, we assume that observations are not identically distributed. Then, we place $x \in [85,120]$ at one single agent and restrict others to $x \in [70,85]$ such that $\AGENTS-1$ agents have statistically insufficient local observations. Moreover, we plot the Mean Squared Error (MSE) of predictions over a test dataset. 

\begin{figure}
\includegraphics[width=0.5\textwidth]{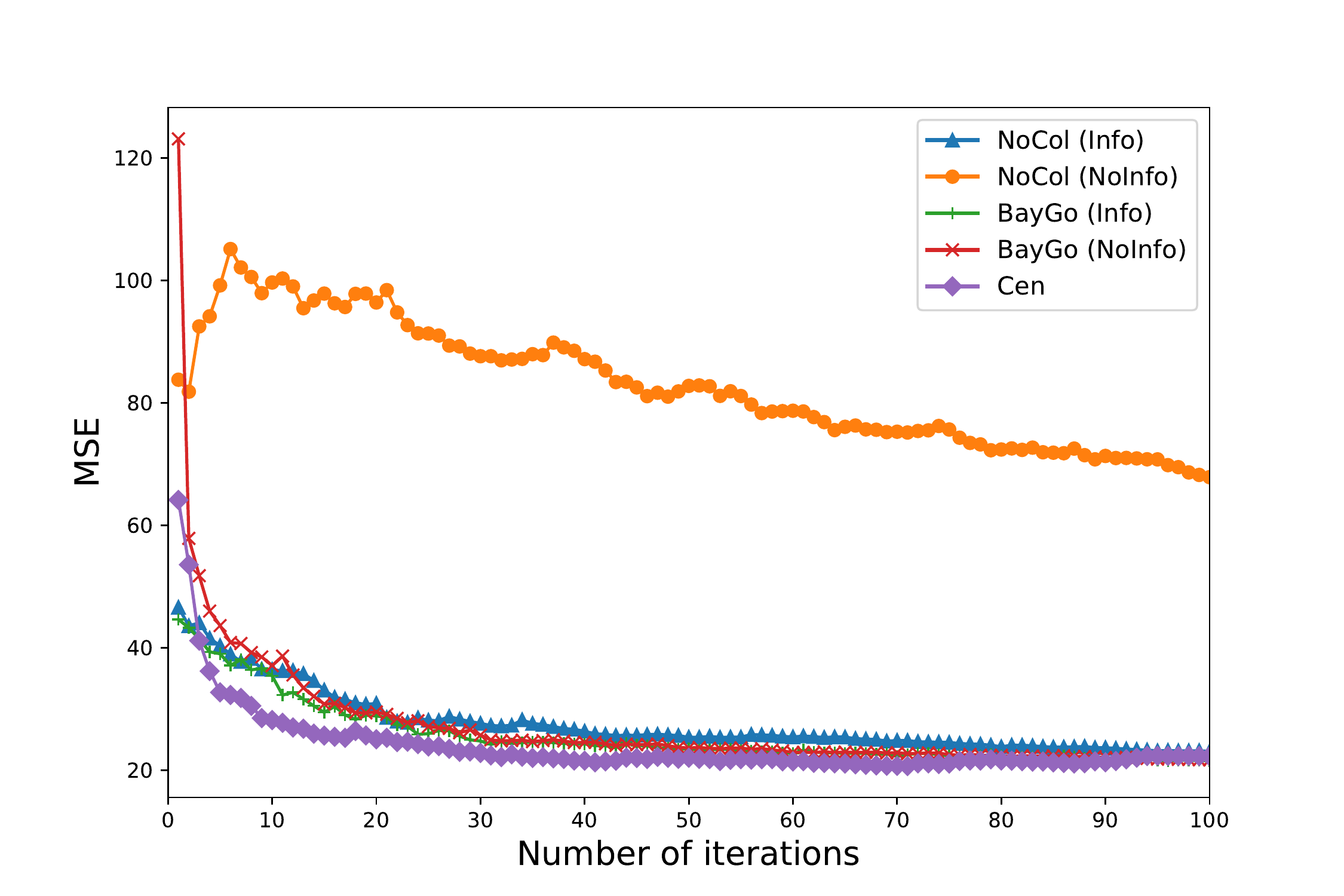}
\centering
\captionof{figure}{Performance comparison between BayGo and benchmarks (Cen and NoCol) in terms of accuracy and convergence speed.}
\label{benchmarks}
\end{figure}
Fig. \ref{benchmarks} shows that both informative (Info) and non informative (NoInfo) agents under our framework achieve comparable performance in terms of MSE and convergence speed compared to the ideal benchmark (Cen) where a single agent has access to all the training data. In contrast to that, in the non-collaborative learning setting (NoCol), the informative agent performs well compared to the single agent case. Nevertheless, non informative agents perform poorly due to their local statistical insufficiency. For ease of presentation, we show the performance of one non informative agent since all non informative agents have comparable performance. 
\begin{figure}
\includegraphics[width=0.5\textwidth]{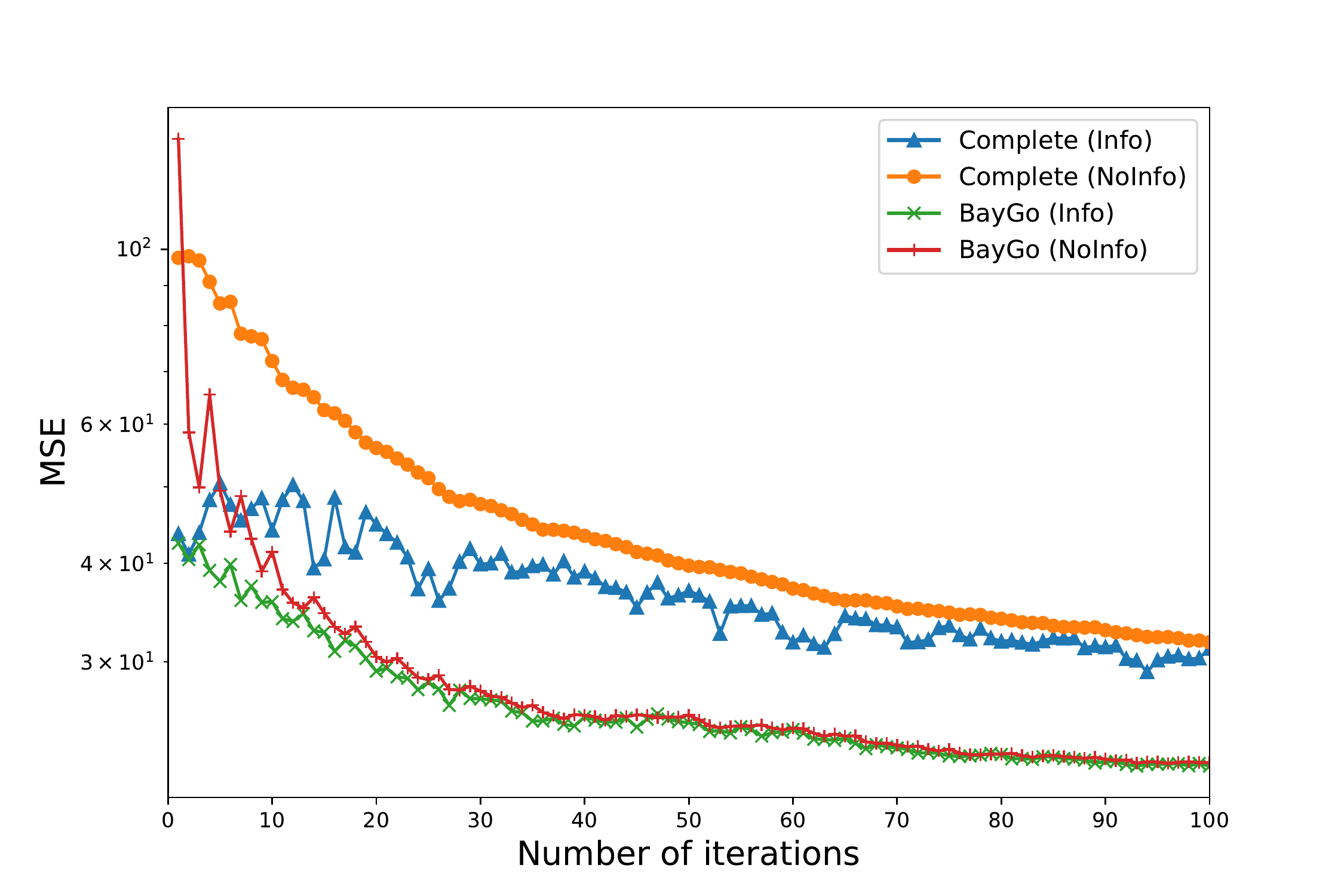}
\centering
\captionof{figure}{Inference accuracy and convergence speed comparison between learning over fully connected graph and BayGo.}
\label{fig:Ng1} 
\end{figure}

In Figure \ref{fig:Ng1}, we compare our approach with the case of learning a global model over a fully-connected graph. As we can observe from the figure, in the case of a fully-connected graph, the informative agent has lower convergence speed due to the poor influence of non informative agents on its learning. i.e., incorporating all non informative agents' updates undermines the learning process of the informative agent. On the contrary, non informative agents benefit from their collaboration with the informative agent, though, this collaboration is not highly utilized since they still incorporate each others' updates which means less mixing weights given to informative agent updates. This can be clearly observed looking into their convergence speed. On the other hand, using our proposed framework, both informative and non informative agents achieve faster convergence. Specifically, we can see that the non informative agents performance matches the informative agent performance after about $20$ iterations without hurting the learning process of the informative agent. In this regard, it is worth mentioning that at time $t=1$, the informative agent in both cases starts from about the same MSE value. Nevertheless, using our approach, the informative agent's MSE value keeps decreasing at each time step unlike the fully-connected graph case where the informative agent performance worsens during the first $20$ iterations before decreasing slowly. 
\begin{figure}
\includegraphics[width=0.5\textwidth]{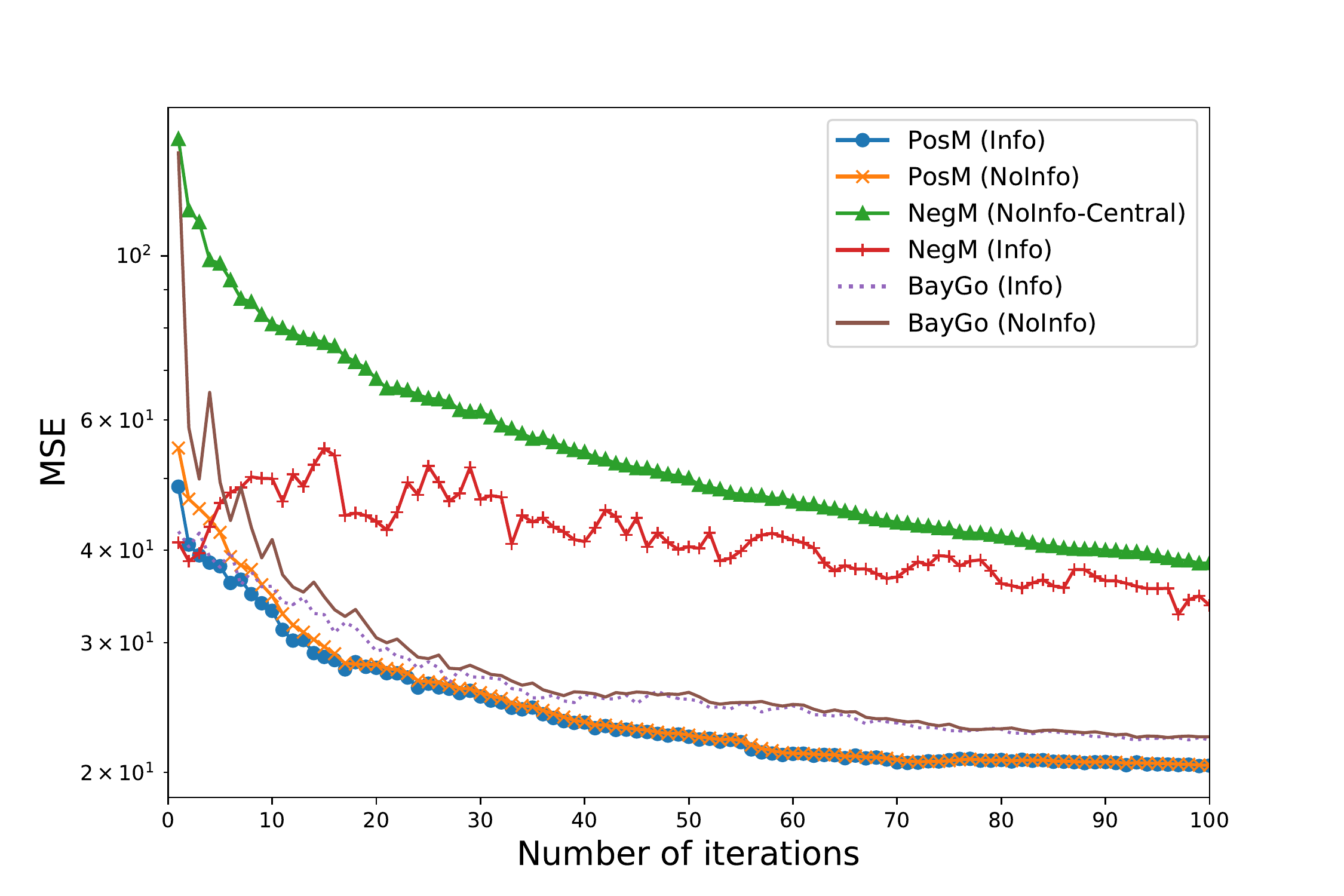}
\centering
\captionof{figure}{Impact of learning using star topology with the informative agent at the center or edge compared to BayGo.}
\label{fig:Ng2}
\end{figure}

In Fig. \ref{fig:Ng2}, we compare our proposed approach with two scenarios based on a star topology. In the first case which we refer to as positive matching (PosM), the most informative agent is placed in the center of the star, and becomes the most influential agent. In the other case which is referred to as negative matching (NegM), the most informative agent is placed at the edge (i.e. not influential). Fig. \ref{fig:Ng2} shows that in the PosM case, the performance of both central and edge agents matches in a very short time and they both perform well in terms of convergence speed and accuracy. On the other hand, under NegM design, we can see a slow convergence for both agents; the informative and the central agents. This is happening because of the large influence of the central non informative agent on the learning of all other agents. Finally, we can see that our approach's performance matches the posM star topology regardless of the fact that agents in our approach have no prior knowledge of informativity states of any of them unlike the pre-assigned posM star topology. 

\begin{figure}
\includegraphics[width=0.5\textwidth]{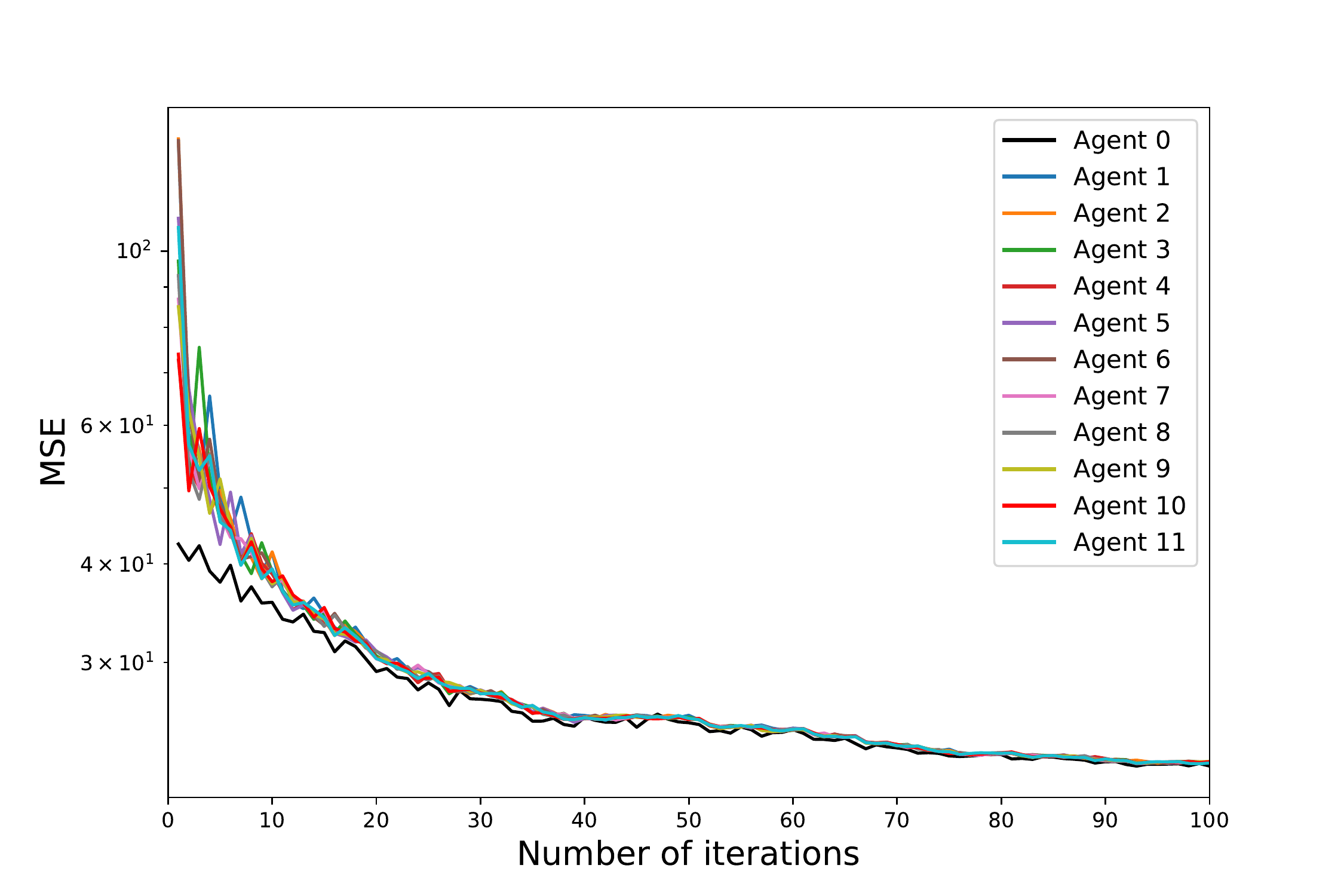}
\centering
\captionof{figure}{MSE of predictions for all agents under our framework.}
\label{consensus}
\end{figure}

Finally, we track the MSE values of all agents in the network in Fig. \ref{consensus} to show that all agents reach consensus in about $20$ iterations which demonstrates that the optimized graph is globally connected since it is a necessary condition for consensus \cite{13}. 
\section{conclusion}
\label{conclusion}
In this paper, we proposed \myFramework{}, a novel fully decentralized joint Bayesian learning and graph optimization framework which ensures fast convergence over a heterogeneous and sparse graph without any assumptions on the prior knowledge of the data distribution across agents. The proposed framework is based on alternating minimization where two subproblems are optimized in an alternating way. We theoretically showed that by optimizing the proposed objective function, the estimation error of the posterior probability decreases exponentially at each iteration. Our extensive simulations show that our framework outperforms the fully-connected and star topology graphs in terms of rate of convergence and learning accuracy. 
\bibliographystyle{IEEEtran}
\bibliography{references}
\end{document}